\documentclass{article} 
\usepackage{iclr2021_conference,times}

\usepackage{amsmath,amsfonts,bm}

\def\eqref#1{equation~\ref{#1}}

\def\1{\bm{1}}

\DeclareMathAlphabet{\mathsfit}{\encodingdefault}{\sfdefault}{m}{sl}
\SetMathAlphabet{\mathsfit}{bold}{\encodingdefault}{\sfdefault}{bx}{n}

\usepackage[colorlinks=true,citecolor=.]{hyperref}
\usepackage{url}

\usepackage{inconsolata}
\usepackage{url}
\usepackage{eurosym}
\usepackage{todonotes}
\usepackage{subcaption}
\usepackage{amssymb}
\usepackage{amsmath}
\usepackage{enumitem}
\usepackage{array}
\usepackage{longtable}
\usepackage{makecell}
\usepackage{xspace}
\usepackage{xcolor,colortbl}
\usepackage{tcolorbox}
\usepackage{arydshln}
\usepackage{adjustbox}

\usepackage{tikz}
\usepackage[tikz]{bclogo}
\usepackage{pgfplots}
\pgfplotsset{width=1.0\columnwidth}
\usepackage{multirow}

\usepackage{makecell}
\usepackage{pifont}
\usepackage{bbm}
\usepackage{rotating}
\usepackage{tablefootnote}
\usepackage{soul}
\usepackage{booktabs}
\usepackage{tikz}
\usepackage[tikz]{bclogo}
\usepackage{pgfplotstable}
\usepackage{mdframed}
\usetikzlibrary{pgfplots.statistics}

\title{From Zero to Hero: Examining the Power of Symbolic Tasks in Instruction Tuning}

\newcommand\sqa{\textsc{SQA}\xspace}
\newcommand\wikisql{\textsc{WikiSQL-Weak}\xspace}
\newcommand\wtq{\textsc{WTQ}\xspace}
\newcommand\tabfact{\textsc{TabFact}\xspace}
\newcommand\scale[1]{{\fontfamily{mathtt}\selectfont {#1}}\xspace}

\definecolor{myred}{hsb}{1, 1., 0.9}
\definecolor{royalblue}{rgb}{0.255, 0.412, 0.882}
\definecolor{crimsonred}{rgb}{0.863, 0.078, 0.235}

\newcommand{\flan}[0]{FLAN-T5\xspace}
\newcommand{\flanlarge}[0]{FLAN-T5 (\scale{Large})\xspace}
\newcommand{\flanxl}[0]{FLAN-T5 (\scale{XL})\xspace}

\newcommand\our{\textcolor{myred}{\textsc{TaPEx Zero}}\xspace}
\newcommand\ourlarge{\textcolor{myred}{\textsc{TaPEx Zero} (\scale{Large})}\xspace}
\newcommand\ourxl{\textcolor{myred}{\textsc{TaPEx Zero} (\scale{XL})}\xspace}
\newcommand\codex{GPT-3 (\scale{code-davinci-002})\xspace}
\newcommand\chatgpt{ChatGPT (\scale{gpt-3.5-turbo})\xspace}

\newcommand{\reffig}[1]{Figure~\ref{#1}}
\newcommand{\refsec}[1]{\S\,\ref{#1}}
\newcommand{\reftab}[1]{Table~\ref{#1}}

\definecolor{quoteborder}{RGB}{234,236,238}
\definecolor{quotebg}{RGB}{246,248,250}

\newmdenv[
  backgroundcolor=red!05,
  linecolor=quoteborder,
  skipabove=1em,
  skipbelow=0em,
  leftline=true,
  topline=false,
  bottomline=false,
  rightline=false,
  linecolor=red!66,
  linewidth=4pt
]{githubquote}

\pgfplotsset{compat=1.17}

\author{Qian Liu\thanks{The first two authors contributed equally. Corresponding to \url{liuqian@sea.com}.} \\ Sea AI Lab
  \And Fan Zhou$^*$ \\ Shanghai Jiao Tong University
  \And \quad\;\;
  \AND Zhengbao Jiang \\ Carnegie Mellon University
  \And Longxu Dou \\ National University of Singapore
  \And Min Lin \\ Sea AI Lab
}

\iclrfinalcopy 
\begin{document}

\maketitle

\begin{abstract}

Fine-tuning language models on tasks with instructions has demonstrated potential in facilitating zero-shot generalization to unseen tasks.
In this paper, we introduce a straightforward yet effective method for enhancing instruction tuning by employing \textit{symbolic tasks}.
Compared to crowdsourced human tasks or model-generated tasks, symbolic tasks present a unique advantage as they can be easily generated in vast quantities, theoretically providing an infinite supply of high-quality training instances.
To explore the potential of symbolic tasks, we carry out an extensive case study on the representative symbolic task of SQL execution.
Empirical results on various benchmarks validate that the integration of SQL execution leads to significant improvements in zero-shot scenarios, particularly in table reasoning.
Notably, our $3$B model surpasses both the $175$B GPT-3 and ChatGPT in zero-shot table reasoning across four benchmarks.
Furthermore, experimental results on BBH ($27$ tasks) and MMLU ($57$ tasks) reveal that language models can be enhanced through symbolic tasks without compromising their generality.
We hope that our paper serves as a catalyst, inspiring increased efforts to incorporate symbolic tasks in instruction tuning.
\end{abstract}

\section{Introduction}

In recent years, the development of large language models (LM) has been one of the most significant advances in natural language processing (NLP)~\citep{devlin-etal-2019-bert,raffel2020exploring,gpt3}.
After being trained on large amounts of natural language corpus~\citep{commoncrawl,pile_dataset} with the language modeling objective, LMs have demonstrated impressive performance on a variety of NLP tasks.
Furthermore, recent progress show that LMs are able to do zero-shot task generalization, which means they can adapt to unseen tasks without any specific fine-tuning on those tasks.
Along this way, one promising direction is the \textit{instruction tuning}~\citep{mishra-etal-2022-cross,flan,t0,instructgpt}.
By fine-tuning LMs to follow instructions on diverse tasks, instruciton tuning enables LMs to perform well on tasks that they have not been explicitly trained on.

Despite the popularity, most existing instruction tuning methods focus on crowdsourced human tasks \citep{flan,mishra-etal-2022-cross,instructgpt} or model-generated tasks \citep{self_instruct} for instruction tuning, which are either in limited quantity or quality.
As scaling up language models in different dimensions has shown promise in pushing the boundaries of zero-shot performance, the search for high-quality and scalable instruction tuning tasks has become increasingly important.
In this paper, we draw inspiration from the work of \textsc{TaPEx}~\citep{liu2021tapex} and investigate the use of \textit{symbolic tasks} as a complementary training resource for instruction tuning.

\begin{figure}[h]
    \centering
    \includegraphics[width=0.95\textwidth]{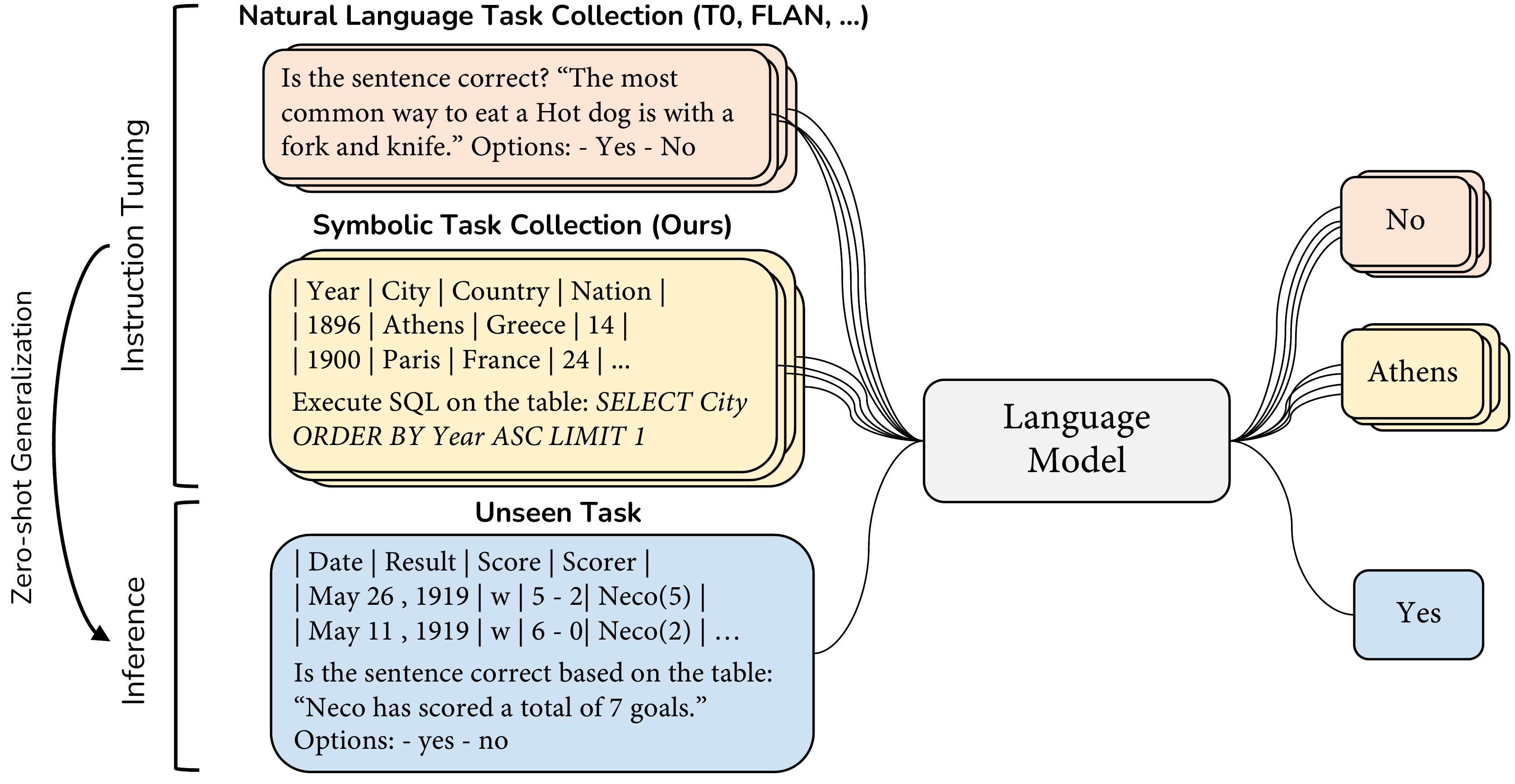}
    \caption{Language models are fine-tuned with both natural language tasks (e.g., fact verificaiton) and symbolic tasks (e.g., SQL execution), and are evaluated on unseen tasks (e.g., table fact verificaiton). }
    \label{fig:model_overview}
    \vspace{-0.7cm}
\end{figure}
A symbolic task is a form of problem-solving or computational process that entails the manipulation of symbolic representations or structured data (e.g., sorting a disordered array).
In general, symbolic tasks are characterized by the use of symbols in formal systems, such as logic, mathematics, or programming languages, instead of relying on natural language.
Unlike natural language, symbols in symbolic tasks follow a limited set of grammatical rules, making them easier to synthesize systematically.
Meanwhile, these symbols can quickly expand in scale through composition, and can be executed on symbolic executors to yield the corresponding output.
As a result, generating diverse, complex, and consistent examples for symbolic tasks is relatively easy compared to crowdsourced human tasks or model-generated tasks.
This ease of generation theoretically allows for a large number of high-quality examples to be created for instruction tuning.
Given the advantageous properties of symbolic tasks, we believe symbolic tasks can serve as a strong complement as a source for instruction tuning.
The core insight is that if a model trained with instruction tuning is capable of performing both symbolic tasks and natural language tasks, then \textbf{the underlying abilities learned from symbolic tasks can be transferred to unseen natural language tasks}.

Motivated by the above, this paper presents a comprehensive study on SQL execution as a representative example to examine the power of symbolic tasks in instruction tuning.
SQL execution refers to the process of executing SQL statements to interact with tables.
Our goal is to extend this interaction between SQL and tables to NL-based table reasoning without using any table reasoning example.
Empirical results demonstrate that the incorporation of SQL execution yields substantial improvements across different benchmarks of table reasoning.
To highlight, our $3$B model outperforms both $175$B GPT-3 (\scale{code-davinci-002}) and ChatGPT (\scale{gpt3.5-turbo}) in zero-shot table reasoning across four benchmarks.
In addition to being used with training, symbolic tasks can also be leveraged by large language models (e.g., GPT-3) without training, as part of their instructions to boost their performance.
For example, without any realistic example in \wtq, SQL execution can boost the GPT-3 performance from $40.4\%$ to $52.9\%$.
Furthermore, we find the task not only enhances performance on table reasoning but also leads to a significant improvement in mathematical reasoning performance as measured by the SVAMP benchmark, increasing the score from $21.4$\% to $26.5$\%.

More importantly, experimental results on BBH ($27$ tasks) and MMLU ($57$ tasks) show that fine-tuning with symbolic tasks does not hurt the model performance on generic held-out tasks, implying that \textbf{we can improve LMs through symbolic tasks without compromising their generality}. 
Based on this observation, we advocate for the research community to expand the scale of symbolic tasks together, pushing the boundaries of zero-shot learning.
As part of our ongoing commitment to collaborative research, we would love to share credits with all contributors who provides any contribution.
For more specific details of our project, please refer to our open-sourced project \url{https://github.com/sail-sg/symbolic-instruction-tuning}.

\section{Related Work}

Our work leverages symbolic tasks in instruction tuning, and thus our work is closely related to instruction tuning. Meanwhile, since symbolic tasks are one typical case of synthetic tasks, we also discuss about the relationship between ours and previous work which uses synthetic tasks in training.

\paragraph{Instruction Tuning}

Instruction tuning is a popular direction in NLP to enable zero-shot generalization on unseen tasks \citep{ye-etal-2021-crossfit,mishra-etal-2022-cross,flan,instructgpt,flanv2,ye2023incontext,wang-etal-2022-super,self_instruct,min-etal-2022-metaicl}.
The core idea behind instruction tuning is that LMs are fine-tuned to accomplish diverse tasks by following a set of instructions. 
Therefore, the task source becomes important in instruction tuning~\citep{flan_collection}.
Previous studies generally collect existing publicly avaiable datasets for the training, including T0~\citep{t0}, FLAN~\citep{flan} and NaturalInstructions~\citep{mishra-etal-2022-cross}, or carefully crowdsource diverse tasks via human annotations, including Super-NaturalInstructions~\citep{wang-etal-2022-super} and InstructGPT~\citep{instructgpt}.
While these mixtures of tasks are generally of high quality, they often rely on a significant amount of human effort and are usually limited in quantity.
We believe that our proposed symbolic tasks could be a valuable addition to address the quantity limitation in this line.
Recently, several researchers have also explored the usage of model-generated tasks, which invoke a large language model (e.g., GPT-3) to generate a diverse set of instructions, task inputs, and task outputs based on a high-quality seed set.
However, these model-generated datasets not only rely on a powerful LM, but also introduce a lot noise (i.e., task output does not correspond to the task input) in the data.\footnote{Some discussion about the quality of GPT-3 generated data in the Alpaca project \citep{alpaca} can be found in \url{https://github.com/gururise/AlpacaDataCleaned}.}
In contrast, in symbolic tasks, we can ensure that the task output corresponds to the task input because the output is automatically obtained from reliable symbolic executors.

\paragraph{Synthetic Training}

Synthetic training is a technique that leverages synthetic tasks, whose examples can be synthesized automatically, to improve the model performance on downstream tasks.
Previous research has explored the power of synthetic tasks in several aspects, including attaining natural language pre-training performance using an artificial task \citep{wu2022insights}, the usage of SQL execution in table pre-training \citep{liu2021tapex,jiang-etal-2022-omnitab}, the application of synchronous context-free grammars in text-to-SQL parsing \citep{zhong-etal-2020-grounded,Yu2020GraPPaGP}, the use of synthetic data in numerical reasoning \citep{geva-etal-2020-injecting,pi-etal-2022-reasoning,trivedi-etal-2022-teaching}, and the use of environment-adaptive synthetic data in language-based environment manipulation tasks \citep{shi-etal-2022-lemon}.
In this line, the most related work to ours is the work of \textsc{TaPEx}~\citep{liu2021tapex}, which introduces SQL execution in the pre-training phase.
However, \textsc{TaPEx} still required fine-tuning on downstream datasets to work~\citep{jiang-etal-2022-omnitab}, while our work can perform zero-shot generalization on downstream tasks without any fine-tuning on the downstream task.
More recently, \citet{li2022unsupervised} also made a noteworthy contribution in the field of zero-shot table reasoning.
Nonetheless, our approach differs in its strategy for leveraging symbolic tasks. 
For example, their method relies on a trained NL generator to ensure naturalness of the model input, which necessitates parallel program and NL corpus.
In contrast, our approach does not require translating the program in symbolic tasks to NL. Instead, we focus on leveraging instruction tuning to directly project the underlying abilities in symbolic tasks onto unseen NL tasks.

\section{Examing the Power of Symbolic Tasks}

In this section, we first describe how we conduct the data synthesis for the SQL execution corpus, and then we introduce two ways to leverage our proposed symbolic tasks in instruction tuing: \textit{multi-task fine-tuning} and \textit{synthetic demonstrations}.
The multi-task fine-tuning method is a training-involved use case of our symbolic tasks, where we need jointly training the LM on the symbolic task and other NL tasks.
The synthetic demonsration method is a training-free use case, where we only need synthesize the symbolic examples and insert them into the current instruction, and then feed them into LMs.
Below we first briefly describe the data synthesis procedure of the SQL execution task, then we describe the two methods in detail.

\subsection{Data Synthesis}

Since we leverage SQL execution as the only symbolic task, the most important factors for the data synthesis are the table source and the SQL queries.
Regarding the source of the table, we go into more details in the next sections, and first we introduce how the SQL query is synthesized.

In principle, we follow \citet{liu2021tapex} to synthesize training instances for SQL execution of symbolic tasks.
Concretely, we obtain them using the SQL templates from the \textsc{Squall} dataset~\citep{shi-etal-2020-potential}.
For instance, a typical SQL template might look like \texttt{\small SELECT num$_1$ WHERE text$_1$ = val$_1$}, with \texttt{\small num$_1$} representing a numeric column and \texttt{\small text$_1$} a text column, while \texttt{\small val$_1$} corresponds to a specific cell value related to the \texttt{\small text$_1$} column. To create concrete SQL queries, we randomly choose headers and cell values from a given table to populate the template.
By instantiating SQL templates on the table source, we can synthesize high-quality examples at any scale.

Given an executable SQL query and a table $T$, where $T$ consists of $M$ rows $\{r_i\}_{i=1}^M$ and $N$ headers $\{c_j\}_{j=1}^N$, we first flatten the table into a sequence $T^*$, and then concatenate it with the SQL query and the task instruction to create each training example.
A typical example can be found in~\reffig{fig:mutli_task_fintune}, where the task instruction is ``{\small \texttt{Execute SQL on the table}}'' and the model is ``{\small \texttt{|Year|City|Country|\,|1896|Athens|Greece|\,|1900|Paris|France|... Execute SQL on the table: SELECT City WHERE Country = Greece ORDER BY Year ASC LIMIT 1}}''.
The supervision for the model is the execution result of the SQL query ``{\small \texttt{Paris}}'', which can be obtained by running the SQL query through an off-the-shelf SQL executor, such as MySQL.

It is worthy to note that the task instruction is always ``{\small \texttt{Execute SQL on the table}}'' for all symbolic task examples.
And the table representations shown in~\reffig{fig:model_overview} and ~\reffig{fig:mutli_task_fintune} are for illustration purpose.
In practise, to obtain the flattend table $T^*$ without losing the table structure information, we include various special tokens to indicate the boundaries inside the table.
Denoting a flattened table as $T^*=\texttt{\small [HEAD]},c_1,{\cdots},c_N,\texttt{\small [ROW]},1,r_1,\texttt{\small [ROW]},2,r_2,{\cdots},r_M$, the special tokens \texttt{\small [HEAD]} and \texttt{\small [ROW]} represent the table header and rows respectively. Additionally, the number following \texttt{\small [ROW]} indicates the row index.

\subsection{Training-Involved Use Case: Multi-Task Fine-tuning}

\begin{figure}[tb]
    \centering
    \includegraphics[width=0.9\textwidth]{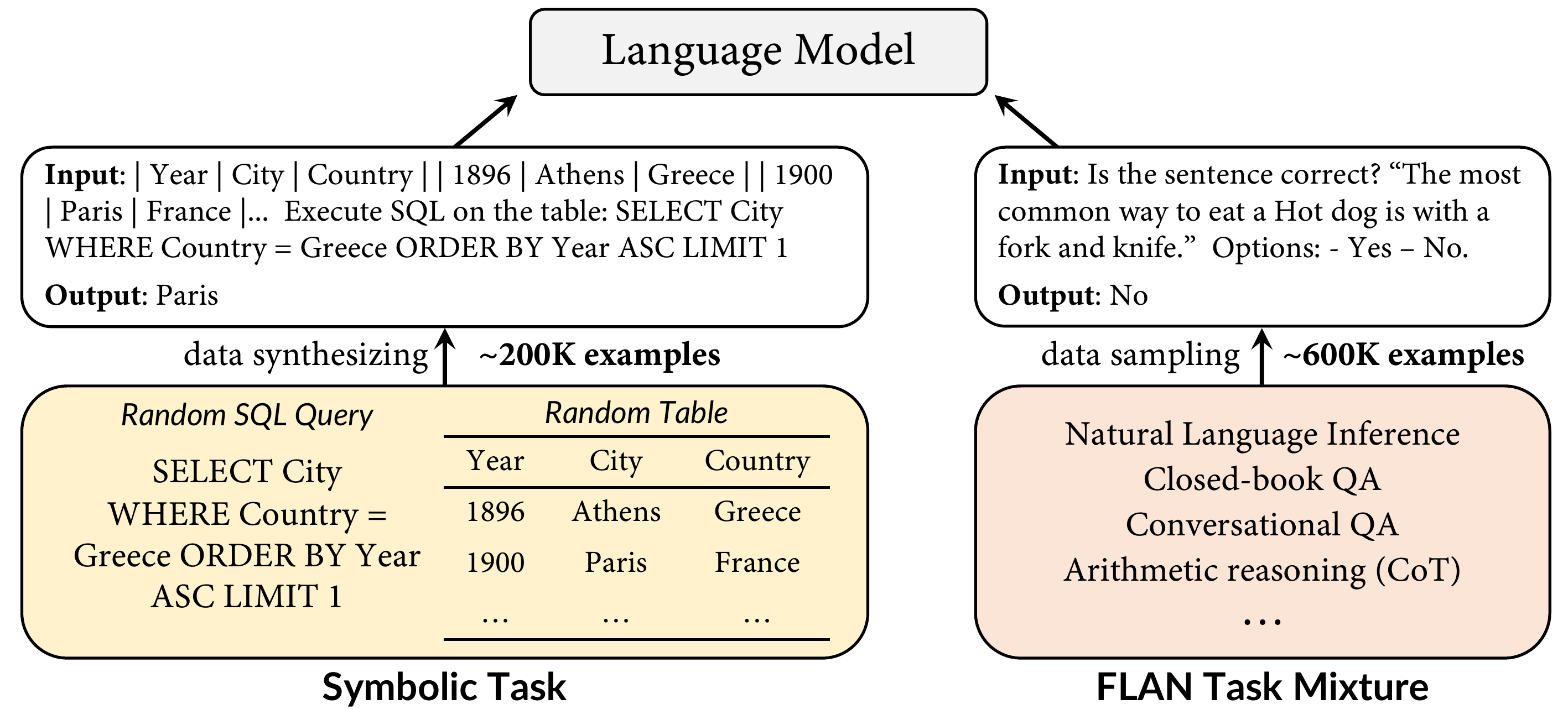}
    \caption{The illustration of our multi-task fine-tuning procedure, where nearly $200$K examples from the symbolic task are merged with $600$K examples from the FLAN task mixture~\citep{flan_collection}. The language model is initialized from FLAN-T5~\citep{flanv2}.}
    \label{fig:mutli_task_fintune}
\end{figure}

Several works have demonstrated the importance of mult-task fine-tuning in the instructon tuning paradigm~\citep{flan,self_instruct}, especially the diversity of tasks.
Motivated by them, the straightforward method to leverage the symbolic task is to incorporate it as part of the fine-tuned tasks, with diverse examples.
To make the task more diverse, we follow~\citet{liu2021tapex} to synthesize SQL queries based on a collection of different tables.
Meanwhile, rather than crawling noisy tables from the Internet and subsequently applying heuristic filtering, we choose clean tables directly from available public datasets.
Specifically, we randomly select close to $1000$ tables from the \wtq training set ~\citep{pasupat2015compositional} to serve as the table source for the SQL execution task.

To verify the proposed approach in multi-task fine-tuning, we follow the popular instruction tuning work FLAN~\citep{flanv2}, which has demostarted great zero-shot performance on several challeging benchmarks such as MMLU~\citep{MMLU} and BBH~\citep{bigbench}.
However, FLAN collected $1836$ NL tasks for instruction tuning, which consumes a lot of time and resources to perform fine-tuning from scratch.
Therefore, inspried by \citet{scialom-etal-2022-fine} which argues that fine-tuned LMs can continually learn new tasks without catastrophic forgetting, we employ the \textit{rehearsal} strategy, where a small amount (e.g., $1$\%) of data in FLAN tasks is replayed during training.
As shown in~\reffig{fig:mutli_task_fintune}, the training starts from the weights of \flan, and we leverage $600$K examples from the FLAN task mixture, along with $200$K examples from our synthetic corpus.
Compared to multi-task fine-tuning from scratch, such kind of rehearsal strategy saves a lot of computation and allows us to reuse the well-trained \flan weights.

\subsection{Training-Free Use Case: Synthetic Demonstration}

\begin{figure}[bt]
    \centering
    \includegraphics[width=0.95\textwidth]{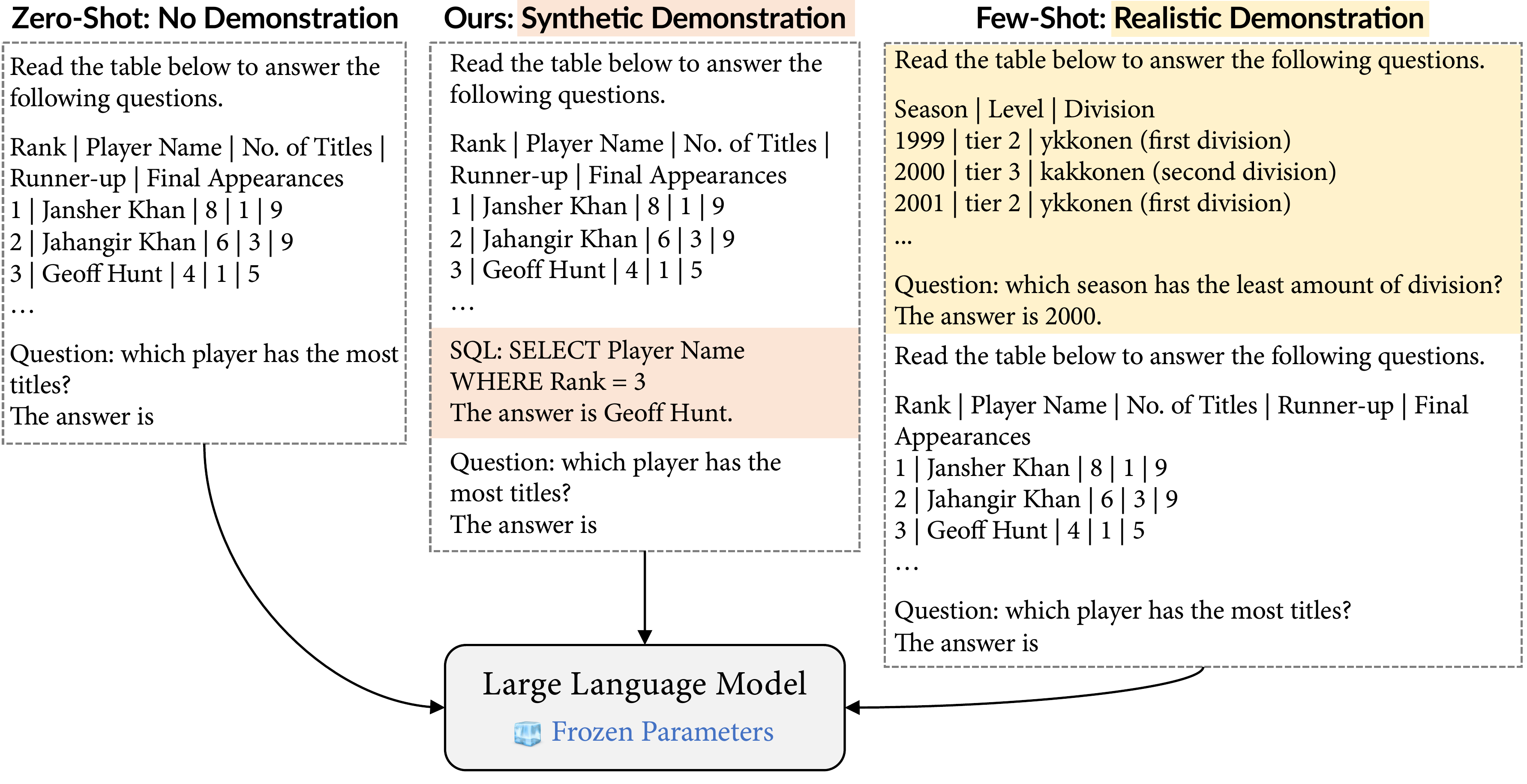}
    \caption{The comparison between different instruction strategies on \wtq.}
    \label{fig:demo_methods}
\end{figure}
Multitasking fine-tuning is applicable to language models in general, but not necessarily to large language models, as the cost of fine-tuning can be high.
To explore whether symbolic tasks work without fine-tuning, we suggest injecting them to LMs by means of synthetic demonstrations, where the symbolic task is employed as part of the instruction to LMs.
\reffig{fig:demo_methods} illustrates different instruction strategies to prompt large LMs, including the \textit{zero-shot}, \textit{synthetic demonstration} and \textit{realistic demonstration}.

Under the zero-shot setting, the instruction to the model only contains the task description and the task input, without any demonstration, while the few-shot setting introduces a few downstream examples as realistic demonstrations for the model to perform in-context learning~\citep{gpt3}. 
Different from both them, we argue that we can leverage the SQL query and its execution result, i.e., the synthetic demonstration, as part of the instruction.
As shown in the middle of~\reffig{fig:demo_methods}, the synthetic demonstrations can be flexibly synthesized without accessing realistic examples.
In practice, we have the flexibility to synthesize the corresponding SQL queries and obtain their execution results based on the task related table.

\section{Experiments}

In this section, we perform experiments to answer three research questions (RQ):
\begin{itemize}[leftmargin=0.4cm]
    \item \textbf{RQ1}. Does the method benefit table reasoning without using any realistic example?
    \item \textbf{RQ2}. Does the method benefit tasks beyond table tasks? 
    \item \textbf{RQ3}. Does the method hurt the model performance on generic tasks? 
\end{itemize}

\subsection{Experimental Setup}

\paragraph{Evaluation Task} We evaluate our approach on $7$ tasks. For \textbf{RQ1}, we perform experiments on typical table reasoning benchmarks, such as weakly-supervised WikiSQL (\wikisql)~\citep{zhongSeq2SQL2017}, WikiTableQuestions (\wtq)~\citep{pasupat2015compositional}, \sqa~\citep{iyyer2017search}, and \tabfact~\citep{chen2019tabfact}.
The first three tasks are in the line of table question answering, while the last one focuses on table fact verification.
\wikisql and \wtq are both large crowd-sourced datasets for developing single-turn question answering systems on tables.
Compared to \wikisql, which necessitates only filtering and optionally aggregating cell values, \wtq demands more advanced reasoning capabilities like sorting.
\sqa is a conversational benchmark, which requires models to take into account the conversational context.
\tabfact is a multi-choice task, which is to determine whether a given statement aligns with the facts presented in a specified table.
For \textbf{RQ2}, we conduct experiments on SVAMP~\citep{patel-etal-2021-nlp}, a challenging diagnostic dataset for probing general numercial reasoning over text. Note that we do not include GSM8K~\citep{gsm8k} since small models, such as our base model \flanlarge  and \flanxl, cannot obtain a reasonable performance. We leave investigation on more benchmarks in the future.
For \textbf{RQ3}, we follow~\citet{flanv2} to evaluate models on two benchmark datasets: BBH ($27$ tasks)~\citep{bigbench} and MMLU ($57$ tasks)~\citep{MMLU}. These benchmarks consist of a set of multiple-choice questions, sourced from a group of humans across various domains. These benchmarks have been recognized as challenging for language models to provide accurate answers.

\begin{table}[t]
    \centering
    \small
    \caption{The instruction used in zero-shot evaluation of \wtq for different models. The instruction for {\codex} is borrowed from~\citet{chen2023fewshot}, while the instruction for {\chatgpt} is inspried by the {Promptify} repo.\protect\footnotemark}
    \label{tab:prompt_for_models}
    \begin{tabular}{lp{9cm}}
    \toprule
        \textbf{Model} & \textbf{Prompt with \{variable\}} \\ \midrule
        \codex & Read the table below to answer the following questions.
        
        \{table\}

        Question: \{question\}
        
The answer is \\ \midrule
        \chatgpt & You are a highly intelligent question answering bot. You take the following Table and Question as input and return the answer from the Table. Retain as much information as needed to answer the question at a later time. Your output format is only ``Answer: Extracted Answer'' form, no other form. And the output should be number or entities, as short as possible, without any explanation.

Table:
\{table\}

Question:
\{question\}

Answer: \\ \midrule
        \flan \;\;/\;\;\our & Here is a table: \{table\}. 
        
        Answer the following question: \{question\} \\ \bottomrule
    \end{tabular}
\end{table}

\paragraph{Baseline Methods} Since we aim to build strong zero-shot models for downstream tasks, we mainly compare the performance of our approach with several zero-shot methods. First, we include the base models \flanlarge and \flanxl. Second, large language models are known to have good zero-shot performance, and \codex has been demonstrated to have strong abiltieis in reasoning~\citep{yifei_better}, so we also compare our performance with it. Meanwhile, we also report the performance of \chatgpt on these datasets. \footnote{The results of GPT-3 and ChatGPT were obtained in March 2023 via the OpenAI API interface.}
It is worthy noting that our claim that Codex and ChatGPT perform zero-shot tasks is based on the assumption that they have not been previously exposed to these datasets.
However, to be honest, we have no way of knowing whether these datasets are already included in their training corpora. Last, we also report the performance of state-of-the-art (SOTA) fine-tuned models on these datasets for reference.
Finally, we reported the performance of \textsc{TaPEx}~\citep{liu2021tapex} to demonstrate that it does not have strong zero-shot generalization ability.
The full results can be found in Appendix~\refsec{appendix:1}.
In the below, we use \our to refer to models trained based on \flan using our proposed approach.

\paragraph{Evaluation} Two evaluation metrics are used in the experiements. The first evaluation metric is \textit{denotation accuracy}, which assesses whether the predicted answer corresponds to the ground-truth answer based on set-level equivalence. It is used in \wtq, \wikisql and \sqa.
The second metric is \textit{accuracy}, which is widely used in multi-classification problems and checks whether the predicted answer is correct. The accuracy metric is used in \tabfact, SVAMP, BBH and MMLU.
Unless explicitly stated, the results presented in the following are obtained via zero-shot prompting on the evaluation task. 
Considering that zero-shot models are sensitive to instructions, we follow the criterion in ~\citet{flan} to report model performance.
Concretely, the model performance on the test set is reported based on the best validation instruction for the target task. 
For \flan and \our, we employ a rigorous approach by manually crafting $10$ distinct instructions for each task and selecting the best instruction based on validation results, ensuring the reliability of the results.
For GPT-3 and ChatGPT, we iteratively refine the instructions using a limited set of samples until the models are able to generate the anticipated outcomes.
The instructions for different methods can be found in ~\reftab{tab:prompt_for_models}.

\paragraph{Implementation Details}

We utilize Transformers~\citep{wolf-etal-2020-transformers} to implement our approach, and leverage DeepSpeed~\citep{deepspeed} to fine-tune models.
The multi-task fine-tuning is conducted using a batch size of $64$ with up to $20,000$ steps, requiring approximately $14$ hours using $8$ A100 GPU cards for \flanxl.
With respect to the sampling strategy from the FLAN collection~\citep{flan_collection}, the normalized distribution of different mixtures CoT\,/\,Muffin\,/\,Natural Instructionsv2\,/\,T0-SF is set as $0.1/0.3/0.3/0.3$.
As for the model optimization, we use the AdamW optimizer~\citep{adamw} and the peak learning rate is set to $3{\times}{10}^{-5}$.
The optimal checkpoint is selected based on the mixed validation set of \wtq and MMLU.
As for \codex and \chatgpt, the most important hyperparameter is the temperature.
We set the hyperparameter temperature as $0$ since we found in practice that $0$ yields the best results. This might explain why the results of \codex in our experiments are much better than those reported by~\citet{chen2023fewshot}, as they used a default temperature of $0.7$.

\footnotetext{\url{https://github.com/promptslab/Promptify}}

\subsection{Experimental Results}

\begin{table}[bt]
\small
\centering
\caption{Zero-shot performance of different models on test sets of various datasets. The fine-tuned SOTA performance are taken from \citet{jiang-etal-2022-omnitab} (\wtq), \citet{liu2021tapex} (\sqa \& \wikisql) and \citet{tabfacthuman} (\tabfact).}
\label{tab:zeroshot_results}
\begin{tabular}{lllll}
\toprule
\textbf{Model} & \textbf{\wtq} & \textbf{\sqa} & \textbf{\wikisql} & \textbf{\tabfact} \\
\midrule
Fine-tuned SOTA & $62.8$ & $74.5$ & $89.5$ & $92.1$ \\
\midrule
\textsc{TaPEx} & $4.1$ & $4.0$ & $21.2$ & -- \\
\codex & $40.4$ & $10.5$ & $55.2$ & $64.1$ \\
\chatgpt & $42.9$ & $13.7$ & $26.1$ & $68.8$ \\
\flanlarge & $30.2$ & $18.9$ & $29.0$ & $59.9$ \\
\ourlarge & $41.9$ {\small $(+11.7)$} & $29.9$ {\small $(+11.0)$} & $62.6$ {\small $(+33.6)$} & $63.9$ {\small $(+4.0)$} \\
\flanxl & $39.5$ & $16.8$ & $38.2$ & $66.3$ \\
\ourxl & ${50.2}$ {\small $(+10.7)$} & ${34.1}$ {\small $(+17.3)$}& ${70.5}$ {\small $(+32.3)$} & ${72.3}$ {\small $(+6.0)$} \\
\bottomrule
\end{tabular}
\end{table}

\begin{figure}[t]
 \centering
 \begin{tikzpicture}[scale=0.85]
\begin{axis}[
    ybar=5pt,
    enlargelimits=0.7,
    legend style={at={(0.5,1.35)},
    anchor=north,legend columns=-1},
    symbolic x coords={0,1,2,5,10},
    xticklabels={$0$ (Zero-shot), $1$, $2$, $5$, $10$},
    xtick=data,
    bar width=.44cm,
    enlarge x limits=0.2,
    width=0.9\textwidth,
    height=0.3\textwidth,
    ymajorgrids=true,
    grid style=dashed,
    every node near coord/.append style={font=\small},
    nodes near coords,
    nodes near coords style={anchor=south,font=\small},
    ymin=35,ymax=50,
    ytick distance=5,
    xlabel={The Number of Demonstrations},
    ylabel={Performance (\%)},
    xlabel near ticks
]
 \addplot[fill=crimsonred!60!white,draw=crimsonred] coordinates {
  (0, 40.4)
  (1, 51.4)
  (2, 52.1)
  (5, 51.9)
  (10, 52.9)
 };
 \addplot [fill=royalblue!60!white,draw=royalblue] coordinates {
  (0, 40.4)
  (1, 51.3)
  (2, 53.4)
  (5, 53.9)
  (10, 54.4)
 };
 \legend{Synthetic Demonstration, Realistic Demonstartion}
 \end{axis}
 \end{tikzpicture}
 \caption{The performance of different instruction strategies with \codex on \wtq. Note that our proposed synthetic demonstration does not leverage any realistic example.}
 \label{fig:syn_demo}
\end{figure}
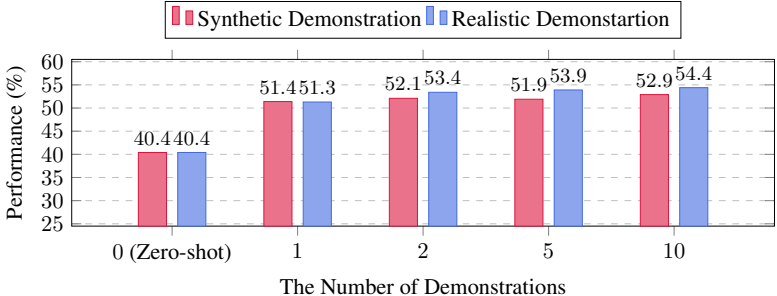

\begin{table}[t]
\small
\centering
\caption{Zero-shot performance of different models on the SVAMP. The fine-tuned SOTA performance is taken from~\citet{pi-etal-2022-reasoning}.}
\label{tab:svamp}
\begin{tabular}{ccccc}
\toprule
{Fine-tuned SOTA} & \flanlarge & {\ourlarge} & \flanxl & {\ourxl} \\
\midrule
$57.4$ & $6.4$ & $8.9$ {\small $(+2.5)$} & $21.4$ & ${26.5}$ {\small $(+5.1)$} \\
\bottomrule
\end{tabular}
\end{table}

We conduct extensive experiments to answer the following research questions:

\begin{githubquote}
\textbf{RQ1}. Does the method benefit table reasoning without using any realistic example? \textbf{Yes}.
\end{githubquote}

\reftab{tab:zeroshot_results} shows the results of different methods on the table reasoning benchmark.
One can observe that GPT-3 and ChatGPT exhibit impressive zero-shot capabilities, such as attaining a denotation accuracy of $42.9$ on the most challenging task \wtq.
Similarly, \flanlarge ($780$M) and \flanxl ($3$B) demonstrate exceptional table reasoning abilities, despite their considerably smaller model scales.
Compared to them, our proposed \our significantly boosts performance based on the \flan models, surpassing all of baseline methods significantly.
Taking the \wtq task as an example, the performance of \ourxl even exceeds $50$\%, which is superior to the fine-tuned performance of \textsc{TaPaS}~\citep{herzig-etal-2020-tapas}, a well-known table pre-training LM.
However, it is worth noting that the performance of all these zero-shot models still falls short compared to the performance of fine-tuned models.

Our approach is not only effective in the multi-task fine-tuning use case, but also shows promise when applied to large language models.
\reffig{fig:syn_demo} compares the performance of our method with realistic demonstrations when paired with GPT-3.
As depicted in the figure, the most significant performance improvement for both synthetic demonstrations and realistic demonstrations comes from the transition from $0$ to $1$ demonsration.
When using only $1$ demonsration, our method achieves a similar performance as realistic demonstration.
As the number of demonstrations gradually increases, the model performance exhibits an upward trend.
Although our approach falls $1.5$\% short of the realistic demonstration when there are $10$ demonstrations, it does not rely on any realistic examples, thereby reducing the need for demonstration tables.
Furthermore, the flexibility of our approach opens the possibility of synthesizing example-specfic demonstrations for different inputs.

\begin{githubquote}
\textbf{RQ2}. Does the method benefit tasks beyond table tasks? \textbf{Yes}.
\end{githubquote}

In addition to table reasoning, our method can also be applied to generic tasks.
Inspired by POET~\citep{pi-etal-2022-reasoning}, we conducted experiments on the SVAMP dataset. 
A typical question in the SVAMP dataset is as ``Jack had 8 pens and Mary had 5 pens. Jack gave 3 pens to Mary. How many pens does Jack have now?''
It is apparent that the question is unrelated to table reasoning, but necessitates numerical reasoning.
\reftab{tab:svamp} compares the performance of \our and \flan on the SVAMP dataset, illustrating a notable improvement of our models over \flan. 
The experimental results indicate that our approach can benefit a wider array of scenarios, not just table reasoning, thereby expanding its potential applications.

\begin{table}[tb]
    \caption{MMLU results using standard few-shot prompting in \flan.}
    \label{tab:mmlu_tab}
    \centering
    \small
    \begin{tabular}{lcccc}
    \toprule
\multirow{2}{*}{MMLU} & \multicolumn{2}{c}{\scale{Large}} & \multicolumn{2}{c}{\scale{XL}} \\
\cmidrule(lr){2-3} \cmidrule(lr){4-5}
& \multicolumn{1}{r}{\flan} & \multicolumn{1}{r}{\our} & \multicolumn{1}{r}{\flan} & \multicolumn{1}{r}{\our} \\ \midrule
    Humanities     & $39.1$          & $44.7$        & $46.3$      & $54.0$     \\
    Social Sciences & $49.1$          & $47.6$        & $57.7$      & $57.4$     \\
    STEM           & $33.2$          & $32.5$        & $39.0$      & $38.4$     \\
    Other          & $47.4$          & $46.2$        & $55.1$      & $51.3$     \\
    \midrule
    Average        & $41.9$          & $41.8$        & $49.3$      & $49.1$     \\
    \bottomrule
    \end{tabular}
\end{table}

\begin{githubquote}
\textbf{RQ3}. Does the method hurt the model performance on generic tasks? \textbf{No}.
\end{githubquote}

Recent studies have shown that models can be made to ``specialize'' in a particular domain via fine-tuning~\citep{model_special}.
However, there seems to be a trade-off between performance in the target domain and in other domains.
For instance, in ~\citet{model_special}, models trained on the GSM8K dataset achieved scores close to zero on the general tasks of BBH and MMLU.

\begin{table*}[t]
\centering
\small
\caption{BBH results using standard few-shot prompting in \flan. We denote algorithmic tasks with superscript $^\S$, and report detailed averages for each sub-task and category. We follow \flan's~\citep{flan2} approach and consider subtasks of two distinct tasks as individual tasks.}
\label{tab:bbh_tab}
\scalebox{1.0}{
\begin{tabular}{lrrrr}
\toprule
\multirow{2}{*}{BIG-bench Hard Task} & \multicolumn{2}{c}{\scale{Large}} & \multicolumn{2}{c}{\scale{XL}} \\
\cmidrule(lr){2-3} \cmidrule(lr){4-5}
& \multicolumn{1}{r}{\flan} & \multicolumn{1}{r}{\our} & \multicolumn{1}{r}{\flan} & \multicolumn{1}{r}{\our} \\ \midrule
Boolean Experssions$^\S$ & $55.6$ & $56.4$ & $56.0$ & $61.2$ \\
Causal Judgement & $57.8$ & $56.7$ & $63.6$ & $59.4$ \\
Date Understanding & $21.6$ & $28.8$ & $44.4$ & $43.6$ \\
Disambiguation QA & $66.0$ & $64.4$ & $67.2$ & $66.0$ \\
Dyck Languages$^\S$ & $1.6$ & $10.0$ & $0.0$ & $3.2$ \\
Formal Fallacies & $54.4$ & $57.2$ & $57.6$ & $56.4$ \\
Geometric Shapes$^\S$ & $20.0$ & $24.8$ & $19.2$ & $24.4$ \\
Hyperbaton & $75.2$ & $64.4$ & $62.4$ & $61.2$ \\
\begin{tabular}[c]{@{}l@{}}Logical Deduction$^\S$ \\ {\small ~~~~~~~~~~~~~(five objects)}\end{tabular} & $43.2$ & $41.2$ & $47.2$ & $50.8$ \\
\begin{tabular}[c]{@{}l@{}}Logical Deduction$^\S$ \\ {\small ~~~~~~~~~~(seven objects)}\end{tabular} & $42.4$ & $45.2$ & $52.4$ & $53.6$ \\
\begin{tabular}[c]{@{}l@{}}Logical Deducction$^\S$ \\ {\small ~~~~~~~~~~~(three objects)}\end{tabular} & $50.8$ & $52.0$ & $60.8$ & $61.6$ \\
Movie Recommendation & $54.0$ & $39.2$ & $56.0$ & $52.8$ \\
Multi-Step Arithmetic$^\S$ & $0.8$ & $0.4$ & $1.6$ & $0.0$ \\
Navigate$^\S$ & $56.0$ & $59.2$ & $60.0$ & $55.2$ \\
Object Counting$^\S$ & $28.4$ & $31.2$ & $36.4$ & $45.6$ \\
Penguins in a Table & $42.5$ & $45.9$ & $40.4$ & $32.9$ \\
Reasoning about Colored Objects & $42.0$ & $39.6$ & $51.6$ & $56.4$ \\
Ruin Names & $21.2$ & $20.0$ & $33.6$ & $35.6$ \\
Salient Translation Error Detection & $40.4$ & $22.8$ & $43.2$ & $42.4$ \\
Snarks & $52.8$ & $54.5$ & $66.3$ & $65.2$ \\
Sports Understanding & $55.2$ & $53.6$ & $57.2$ & $54.4$ \\
Temporal Sequences$^\S$ & $22.8$ & $26.4$ & $21.6$ & $19.6$ \\
\begin{tabular}[c]{@{}l@{}}Tracking Shuffled Objects$^\S$ \\ {\small ~~~~~~~~~~~~~\quad\quad\quad(five objects)}\end{tabular} & $12.4$ & $12.4$ & $12.4$ & $12.4$ \\
\begin{tabular}[c]{@{}l@{}}Tracking Shuffled Objects$^\S$ \\ {\small ~~~~~~~~~~\quad\quad\quad(seven objects)}\end{tabular} & $8.4$ & $8.4$ & $8.4$ & $8.4$ \\
\begin{tabular}[c]{@{}l@{}}Tracking Shuffled Objects$^\S$ \\ {\small ~~~~~~~~~~~\quad\quad\quad(three objects)}\end{tabular} & $33.6$ & $33.6$ & $32.8$ & $33.2$ \\
Web of Lies$^\S$ & $52.8$ & $51.2$ & $51.6$ & $46.0$ \\
Word Sorting$^\S$ & $3.2$ & $0.0$ & $2.4$ & $2.0$ \\ \midrule
NLP Task (\textit{avg}) & $48.6$ & $45.6$ & $53.6$ & $52.2$ \\
Algorithm Task$^\S$ (\textit{avg}) & $30.8$ & $32.3$ & $32.9$ & $34.1$ \\
All Tasks (\textit{avg}) & $37.6$ & $37.0$ & $41.0$ & $40.9$ \\
\bottomrule
\end{tabular}
}
\end{table*}

To investigate whether the performance degration occurs in our approach, we report the performance of \our on the MMLU (\reftab{tab:mmlu_tab}) and BBH (\reftab{tab:bbh_tab}) test sets. 
Notably, following \flan, we incorporate few-shot demonstrations in the instruction and adopt the standard few-shot templates, as outlined in~\citet{flanv2}.
As shown in the table, the performance of \our and \flan on these two datasets is comparable. 
For example, \ourxl can achieve an accuracy of $40.9\%$ on BBH tasks.
Interestingly, our approach demonstrates a greater advantage in specific task categories, such as Humanities in the MMLU dataset and Algorithm tasks in the BBH dataset.
These experimental results demonsrate that our proposed symbolic tasks does not hurt the model performance on generic held-out tasks, implying that we can always improve LMs through symbolic tasks without sacrificing their generality.

\section{Conclusion}

In this paper, we presented a novel approach for instruction tuning that leverages synthetic tasks to improve the performance of language models on unseen tasks. Our proposed approach of utilizing symbolic tasks for instruction tuning has demonstrated its effectiveness in enhancing the performance of language models on unseen tasks, as exemplified by our case study on zero-shot table reasoning. The experimental results on several benchmarks show that our approach outperforms larger language models signifantly and achieves the best zero-shot performance across table reasoning benchmarks.
Meanwhile, the model maintains the original performance on generic datasets, which indicates that we can import symbolic tasks into instruction tuning without sacrificing the generality.
Compared to crowdsourced tasks, symbolic tasks are inherently more controllable, thus offering a number of potential opportunities. In the future, we hope to collaborate with the community to discover more valuable symbolic tasks, ultimately contributing to the development of mainstream instruction fine-tuning models.

\bibliography{custom,anthology}
\bibliographystyle{iclr2021_conference}

\appendix

\newpage
\section{Experimental Results on Tabular Reasoning}\label{appendix:1}

\begin{table}[htbp]
    \small
     \centering
    \begin{minipage}{0.45\linewidth}
    \caption{Denotation accuracies of different models on \wtq.}
    \label{tab:wtq_results}
    \centering
    \scalebox{0.91}{
    \begin{tabular}{clc}
    \toprule
    \textbf{Setting} & \textbf{Model} & \textbf{Test} \\
    \midrule
     \multirow{14}{*}{\rotatebox[origin=c]{90}{Supervised Training}} & \multicolumn{2}{c}{\textit{Previous Systems}} \\
    & \citet{pasupat2015compositional} & $37.1$ \\
    & \citet{DBLP:journals/corr/NeelakantanLS15} & $34.2$ \\
    & \citet{zhang-etal-2017-macro} & $43.7$ \\
    & \citet{liang18mapo}  & $43.8$ \\ 
    & \citet{Dasigi2019IterativeSF} & $44.3$ \\
    & \citet{Agarwal2019LearningTG} & $44.1$ \\ 
    & \citet{wang-etal-2019-learning} & $44.5$ \\
    & \multicolumn{2}{c}{\textit{Language Models}} \\
    & \citet{herzig-etal-2020-tapas} & $48.8$ \\
    & \citet{yin-etal-2020-tabert} & $52.3$ \\
    & \citet{Yu2020GraPPaGP} & $52.7$ \\
    & \citet{liu2021tapex} & $60.1$ \\
    & \citet{zhou-etal-2022-tacube} & $61.3$ \\
    & \citet{jiang-etal-2022-omnitab} & ${62.8}$ \\
        \midrule
    \multirow{6}{*}{\rotatebox[origin=c]{90}{Zero-Shot}}
    & \codex & $40.4$ \\
    & \chatgpt & $42.9$ \\
    & \flanlarge & $30.2$ \\
    & \ourlarge & $41.9$ \\
    & \flanxl & $39.5$ \\
    & \ourxl & ${50.2}$ \\
    \bottomrule
    \end{tabular}
    }
    \end{minipage}
    \hspace{4pt}
    \begin{minipage}{0.5\linewidth}
    \centering
    \caption{Denotation accuracies of different models on \sqa.}
    \label{tab:sqa_results}
    \scalebox{0.95}{
    \begin{tabular}{clc}
    \toprule
    \textbf{Setting} & \textbf{Model} & \textbf{Test (All)} \\
    \midrule
     \multirow{9}{*}{\rotatebox[origin=c]{90}{Supervised Training}} & \multicolumn{2}{c}{\textit{Previous Systems}} \\
        & \citet{pasupat2015compositional}       &  $33.2$\\
        & \citet{DBLP:conf/iclr/NeelakantanLAMA17}       &  $40.2$ \\
        & \citet{iyyer2017search}          &  $44.7$\\
        & \citet{Sun2019KnowledgeAwareCS}     &  $45.6$ \\ 
        & \citet{mueller-etal-2019-answering}     &  $55.1$ \\
        & \multicolumn{2}{c}{\textit{Language Models}} \\
        & \citet{Yu2021SCoRePF} & $65.4$ \\
        & \citet{herzig-etal-2020-tapas} & $67.2$ \\
        & \citet{eisenschlos-etal-2020-understanding} & $71.0$ \\
        & \citet{liu2021tapex} & ${74.5}$ \\
        \midrule
    \multirow{6}{*}{\rotatebox[origin=c]{90}{Zero-Shot}}
    & \codex & $10.5$ \\
    & \chatgpt & $13.7$ \\
    & \flanlarge & $18.9$ \\
    & \ourlarge & $29.9$ \\
    & \flanxl & $16.8$ \\
    & \ourxl & ${34.1}$ \\
    \bottomrule
    \end{tabular}
    }
    \end{minipage}
\end{table}

\begin{table}[htbp]
    \small
    \centering
    \begin{minipage}{0.46\linewidth}
    \caption{Denotation accuracies of different models on \wikisql.}
    \label{tab:wikisql_results}
    \begin{tabular}{clc}
    \toprule
    \textbf{Setting} & \textbf{Model} & \textbf{Test} \\
    \midrule
     \multirow{9}{*}{\rotatebox[origin=c]{90}{Supervised Training}} & \multicolumn{2}{c}{\textit{Previous Systems}} \\
    & \citet{liang18mapo}       &  $72.4$ \\
    & \citet{Agarwal2019LearningTG}  &  $74.8$ \\
    & \citet{wang-etal-2019-learning}     &  $79.3$ \\ 
    & \citet{min-etal-2019-discrete}     &  $83.9$ \\
    & \multicolumn{2}{c}{\textit{Language Models}} \\
    & \citet{herzig-etal-2020-tapas} & $83.6$ \\
    & \citet{Yu2020GraPPaGP} & $84.7$ \\
    & \citet{jiang-etal-2022-omnitab} & $88.1$ \\
    & \citet{liu2021tapex} & ${89.5}$ \\
    \midrule
    \multirow{7}{*}{\rotatebox[origin=c]{90}{Zero-Shot}}
    & \chatgpt & $26.1$ \\
    & \codex &  $55.2$ \\
    & \citet{li2022unsupervised} & $61.6$ \\
    & \flanlarge & $29.0$ \\
    & \ourlarge & ${62.6}$ \\
    & \flanxl & $38.2$ \\
    & \ourxl & ${70.5}$ \\
    \bottomrule
    \end{tabular}
    \end{minipage}
    \hspace{4pt}
    \begin{minipage}{0.45\linewidth}
    \vspace{-0.4cm}
    \caption{Accuracies of different models on \tabfact.}
    \label{tab:tabfact_results}
    \begin{tabular}{clc}
    \toprule
    \textbf{Setting} & \textbf{Model} & \textbf{Test} \\
    \midrule
     \multirow{8}{*}{\rotatebox[origin=c]{90}{Supervised Training}} & \multicolumn{2}{c}{\textit{Language Models}} \\
       & {\citet{chen2019tabfact}}  & $66.1$  \\ 
       & {\citet{zhong-etal-2020-logicalfactchecker}}  & $71.8$ \\
       & {\citet{shi-etal-2020-learn}}  & $72.5$ \\
       & {\citet{zhang-etal-2020-table}} & $73.3$ \\
       & {\citet{yang-etal-2020-program}} & $74.9$ \\
       & {\citet{eisenschlos-etal-2020-understanding}}  & $81.0$ \\
       & {\citet{zhou-etal-2022-table}}  & $85.1$ \\
       & {\citet{gu-etal-2022-pasta}}  & $89.3$ \\
       & {\citet{tabfacthuman}} & $93.0$ \\
    \midrule
    \multirow{6}{*}{\rotatebox[origin=c]{90}{Zero-Shot}}
    & \codex & $64.1$ \\
    & \chatgpt & $68.8$ \\
    & \flanlarge (\scale{Large}) & $59.9$ \\
    & \ourlarge & $63.9$ \\
    & \flanxl (\scale{XL}) & $66.3$ \\
    & \ourxl &  ${72.3}$ \\
    \bottomrule
    \end{tabular}
    \end{minipage}
\end{table}

\end{document}